\documentclass{article}
\usepackage{amssymb}
\usepackage{todonotes}
\usepackage{amsmath}
\usepackage{booktabs}
\usepackage{float}
\usepackage{svg}
\usepackage{subfig}
\usepackage{authblk}

\newtheorem{theorem}{Theorem}[section]

\newtheorem{proposition}[theorem]{Proposition}
\newtheorem{assumption}[theorem]{Assumption}
\usepackage{multirow} 

\usepackage[preprint]{corl_2025} 

\title{Sequence Modeling for Time-Optimal Quadrotor Trajectory Optimization with Sampling-based Robustness Analysis}

\author[1]{Katherine Mao}
\author[2]{Hongzhan Yu}
\author[2]{Ruipeng Zhang}
\author[1]{Igor Spasojevic}
\author[1]{\\M Ani Hsieh}
\author[2]{Sicun Gao}
\author[1]{Vijay Kumar}

\affil[1]{University of Pennsylvania}
\affil[2]{University of California San Diego}

\begin{document}

\maketitle

\begin{abstract}
Time-optimal trajectories drive quadrotors to their dynamic limits, 
but computing such trajectories involves solving non-convex problems via iterative nonlinear optimization, 
making them prohibitively costly for real-time applications. 
In this work, we investigate learning-based models that imitate a model-based time-optimal trajectory planner to accelerate trajectory generation. 
Given a dataset of collision-free geometric paths, we show that modeling architectures can effectively learn the patterns underlying time-optimal trajectories.
We introduce a quantitative framework to analyze
local analytic properties of the learned models,
and link them to the Backward Reachable Tube of the geometric tracking controller. 
To enhance robustness, we propose a data augmentation scheme that applies random perturbations to the input paths.
Compared to classical planners,
our method achieves substantial speedups,
and we validate its real-time feasibility on a hardware quadrotor platform.
Experiments demonstrate that the learned models generalize to previously unseen path lengths.
The code for our approach can be found here: \href{https://github.com/maokat12/lbTOPPQuad}{https://github.com/maokat12/lbTOPPQuad}
\end{abstract}


\section{Introduction}

Optimal trajectory generation is one of the core components of an agile micro aerial vehicle's (MAVs) autonomy stack. 
Numerous applications such as search and rescue operations, disaster response, and package and aid delivery require these robots to perform tasks safely, at operational speeds.  
The aim of minimizing task completion time arises not only due to the limited battery life onboard the MAVs, but also due to the desire to take advantage of their full flight envelope. 

The key algorithmic challenge behind optimization problems underlying synthesizing time-optimal trajectories lies in the non-convexity.
Part of the non-convex constraints come from the presence of obstacles in the environment. 
Another source of non-convexity is the nonlinear nature of robot dynamics. 
The majority of previous approaches have either optimized trajectories by using simplified dynamics models, faithful dynamics models with proxy actuation constraints, or a combination of the two. 
Yet another set of approaches planned trajectories with both faithful dynamics models and suitable actuation constraints. 
Such planners have exhibited superior mission execution time, at the cost of much higher computational resources.

This is the first work to develop a learning-based algorithm for computationally efficient time optimal path parametrization for quadrotors with faithful dynamics and actuation constraints. 
The approach for learning a solution to the high-dimensional sequential optimization problem rests on a combination of domain-specific insights as well as a novel imitation learning formulation for ``sequence-to-sequence'' problems. 

In summary, the contributions of this paper are as follows:
\begin{itemize}
    \item A learning-based formulation for imitating a model-based time-optimal trajectory planner, employing an input–output feature design that predicts only the minimal variables necessary to reconstruct the time-optimal trajectory.
    \item A rigorous robustness analysis framework that quantifies how well predicted trajectories can be tracked by a controller via a sampling-based approach to recover Backward Reachable Tubes (BRT), 
    along with a data augmentation strategy to enhance model robustness.
    \item A comprehensive ablation study across various neural architectures, showing that an LSTM encoder-decoder model achieves near-time-optimal performance with significant speedup over optimized-based planners.
    \item A demonstration of our learning-based planner on a hardware platform,
    showing robust generalization to unseen path geometries and lengths.
\end{itemize}

\begin{figure}[t]
    \centering
    \includegraphics[width=1\linewidth]{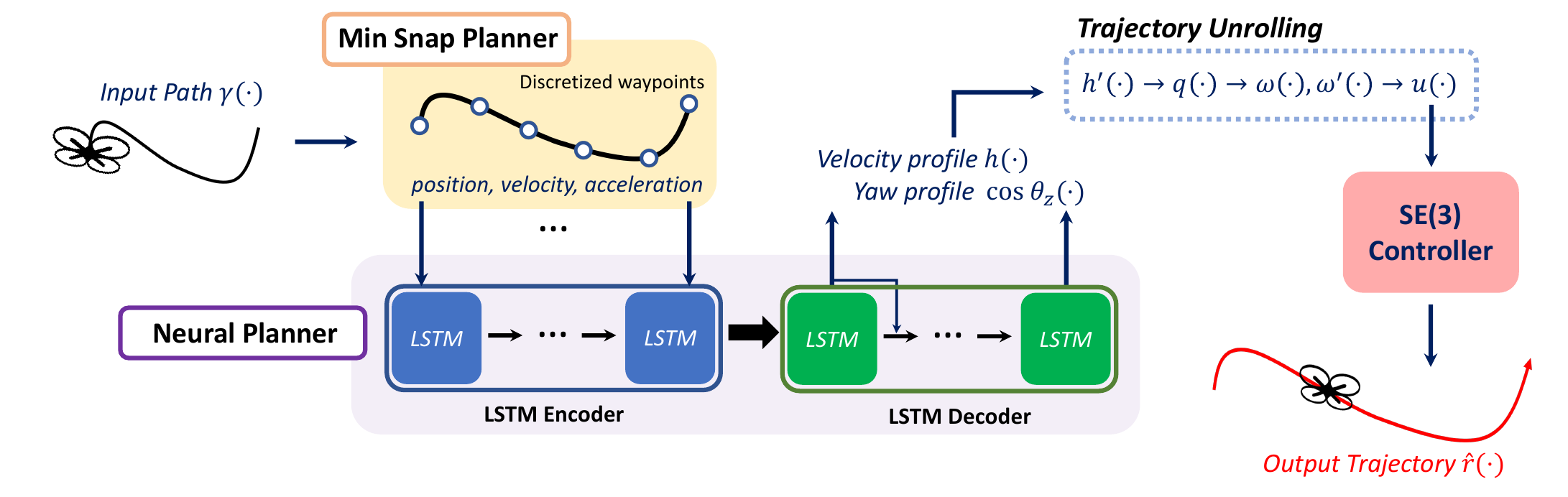}
    \caption{\small 
    We uniformly discretize a geometric path $\gamma(\cdot)$ from a sequence of waypoints using the minimum snap planner \cite{5980409}.
    Based on this discretized trajectory, 
    our LSTM encoder-decoder model (which performed the best in our ablation study) predicts squared-speed $h(\cdot)$ and yaw $\theta_{z}$ profiles,
    imitating the model-based time-optimal planner TOPPQuad~\cite{10801611}.
    From these predicted profiles,
    we unroll the full robot state trajectory.
    Finally, a low-level geometric controller computes the control trajectory to execute, 
    ensuring that per-motor actuation constraints must be satisfied. 
    }
    \label{fig:pipeline}
    \vspace{-10pt}
\end{figure}

\section{Related Work}



The nature of time-optimal quadrotor trajectories requires plans that push the system to its physical limits. Paired with the quadrotor's underactuated dynamics, this remains a challenging problem. 
Popularized by \cite{5980409}, many approaches plan trajectories following a polynomial structure, where trajectories are characterized by a set of waypoints and their time allocations \cite{9765821}\cite{10591215}\cite{7487284}\cite{8264768}. These approaches take advantage of the quadrotor's differential flatness property to ensure smoothness of state values and often allow for faster compute times, but sacrifice the 'bang-bang' behavior required for more aggressive flight. Other approaches forgo the polynomial structure to plan around the full dynamics. A series of approaches track progress along a nominal path, allowing deviations \cite{doi:10.1126/scirobotics.abh1221}\cite{9802523}, while \cite{10801611} determines the time paramterization of a fixed geometric path. Finally, \cite{10700666}, \cite{10342270} simplify the quadrotor to a point model in the planning phase, then rely on modern nonlinear Motion Predictive Controllers (NMPC) to track a potentially dynamically infeasible reference trajectory.


The heavy computation cost of optimization-based time-optimal trajectory planners has made learning-based solutions an attractive option. Although polynomial planners for a set of waypoints can be formulated as a convex problem, the selection of the intermediary waypoints and their corresponding time allocations is still computationally challenging. Approaches to the time allocation problem for a set of waypoints have used Gaussian Processes \cite{ryou2021multi}, sequence-to-sequence learning \cite{ryou2023real}, GNNs~\cite{zhang2022learning} and transformers \cite{pmlr-v211-tankasala23a}. \cite{wu2024deep} broadens this problem and trains an LSTM to learn both the intermediary waypoints and time allocation given a series of safe flight corridors and goal points. While often computationally faster, the lack of full quadrotor dynamic constraints in the polynomial approaches can lead to dynamically infeasible motor thrusts. On the other hand, planners that utilize the full dynamics can become intractably slow, ranging from seconds \cite{10801611} to hours \cite{doi:10.1126/scirobotics.abh1221}. \cite{9636053}, \cite{kaufmann2023champion} solve this by utilizing Reinforcement Learning to generate near-time optimal trajectories for drone racing tracks. However, the progress-based nature of these formulations encourages deviations from a nominal center-line path and cannot guarantee safety in cluttered environments.


\section{Preliminary}

TOPPQuad \cite{10801611} is an optimization-based approach for Time-Optimal Path Parameterization (TOPP) which generates dynamically feasible quadrotor trajectories while ensuring strict adherence to a collision-free path. The key to this method is the squared-speed profile, $h(\cdot)$, which dictates the relationship between traversal time and the progress along a $N$-point discretized geometric path $\gamma(\cdot)$. However, due to the non-convexity of the full dynamic model and the desire for explicit bounds on actuation constraints, the optimization must be performed upon an $16 \times N$ variable state space, incurring heavy computation cost and slow runtime. This state space includes the rotational profile along the trajectory. Conversely, \cite{5980409} shows that a quadrotor is a differentially flat system, where any trajectory can be uniquely represented by four flat variables and their derivatives: position ($x, y, z$) and yaw ($\theta_z$). Given the nature of the TOPP problem and the relationship between $h$ and the higher positional derivatives, the time optimal trajectory along a given path $\gamma(\cdot)$ can be represented as function of just $h$ and $\theta_z$. 

\section{Methodolgy}

\subsection{Imitation Learning Problem Formulation}

When designing an imitation learning framework, 
it is essential to ensure that the input-output mapping is feasible with respect to system constraints.
Furthermore, constraining the output dimensionality to only the minimal set of required features
mitigates overfitting risks and promotes robustness.
Given $\gamma(\cdot)$, TOPPQuad produces a time-optimal, dynamically feasible trajectory $r(\cdot)$, where dynamically feasibility is defined as respecting all state and input constraints.
However, the variables are tightly coupled by the underlying dynamic constraints,
making it challenging to learn them jointly in a direct manner.
In particular, the motor thrust $u(\cdot)$ must lie within specified actuation limits, highly non-linear functions of the flat outputs, 
and the quaternions $q(\cdot)$ must reside on the $3$-sphere space, $S^3 = \{ q \in \mathbb{R}^4 \ \vert \ || q||_2 = 1\}$.

We propose to use $[h(\cdot), \cos\theta_{z}(\cdot)]$ as the output variables, 
where $\cos\theta_{z}(\cdot)$ encodes the yaw rotation 
enconded via the cosine function, in order to address the many-to-one yaw wraparound.  
Next, we discuss how to recover the original variables,
providing equations in Appendix.
We obtain the speed profile derivative $h^{'}$ from finite differences of the learned squared-speed profile $h$.
To construct quaternion $q$, we first compute the body $z$-axis vector $b_{3}$ by adding gravity to the derived acceleration 
(from speed profiles and path curvature)
and normalizing,
ensuring $b_{3}$ aligns with the net thrust vector.
Then, $q$ is derived via rotation composition,
orienting the drone to align its body-$z$ axis with $b_{3}$ and setting yaw to the desired $\theta_{z}$.
Next, we calculate the rotation change between steps,  
yielding angular velocity $\omega$ and its derivative $\omega{'}$.
Finally, a low-level geometric 
controller \cite{se3control}\cite{watterson2019control} computes $u(\cdot)$ from the derived states.
In TOPPQuad, $u(\cdot)$ is jointly optimized with other decision variables to guarantee dynamic feasibility.
Conversely, our approach recovers $u(\cdot)$ from the orientation and accelerations,
ensuring that the per-motor actuation constraints are satisfied.

We augment the model's input with path curvature, 
the first and second derivatives of the geometric path $\gamma{'}(\cdot)$ and $\gamma{''}(\cdot)$.
This directly provides explicit geometric information, 
rather than requiring the model to infer it implicitly.
In summary, we formulate the imitation task as learning the mapping:
\begin{align}
    [\gamma(\cdot), \gamma'(\cdot), \gamma''(\cdot)] \rightarrow [h(\cdot), \cos\theta_{z}(\cdot)],
\end{align}
the minimal set of outputs necessary to reconstruct the time-optimal path parameterization.

\subsection{Robustness Analysis}
\label{sec:robustness_analysis}

To ensure effective generalization, it is crucial to thoroughly evaluate the robustness of the learned models
beyond 
empirical performance metrics. 
Given a geometric path $\gamma(\cdot)$, 
the model predicts $[h(\cdot), \cos\theta_{z}(\cdot)]$ from which we derive the robot state trajectories $r(\cdot) := \{h(\cdot), q(\cdot), \omega(\cdot)\}$.
We assume that the quadrotor starts at $\gamma(s_{1})$ at rest (zero velocity). Let $\hat{r}(\cdot)$ denote the full robot state trajectory after executing the error-tracking low-level controller $U$~\cite{watterson2019control}.
That is, $\hat{r}(s_{1}) = r(s_{1})$, and for each $ i \in \{1, ..., N-1\}$:
\begin{align}
    \hat{r}(s_{i+1}) = \hat{r}(s_{i}) + \int_{0}^{\Delta s} \dot{f}(\hat{r}(s_{i} + \mathrm{d}s), U\Big(\hat{r}(s_{i} + \mathrm{d}s), r(s_{i})\Big)) \mathrm{d}s,
\end{align}
which integrates the quadrotor dynamics $\dot{f}$ over the spatial step-size $\Delta s$.
To formalize robustness, we introduce the concept of finite-time Backward Reachable Tube (BRT)~\cite{bansal2017hamilton}: 

Let $\xi_{U}(x, \Delta t)$ denote the subset of the state space from which state $x$ can be reached within $\Delta t$ seconds under a given $U$. 
Formally,
\begin{align}
    \xi_{U}(x, \Delta t) = \{ x_{0} | \exists \tau \leq \Delta t, \text{ s.t. } x_{0}(\tau) = x \text{ under } U\}.
\end{align}
\begin{proposition} \label{lemma:robustness_no_perturbation}
    Suppose $\gamma(\cdot)$ is a geometric path, and let $r(\cdot)$ and $\hat{r}(\cdot)$ be the planned and the simulated trajectories, respectively. 
    If, for each $i \in \{1, ..., N-1\}$,
    \begin{align}
        \hat{r}(s_{i}) \in \xi_{U}(r(s_{i+1}), t_{i}), \text{ where } t_{i} = 2\Delta s/(\sqrt{h(s_{i})}+ \sqrt{h(s_{i+1})}),
    \end{align}
    then the trajectory planner is robust with respect to dynamic feasibility.
\end{proposition}
Intuitively, if the model perfectly predicts a feasible trajectory, 
we have $r(\cdot) = \hat{r}(\cdot)$.
Proposition~\ref{lemma:robustness_no_perturbation} captures how violations in the planned state are recoverable under the low-level controller $U$.
Even more, if $\hat{r}(\cdot)$ coincides with $\gamma(\cdot)$, 
then the tracking problem is solved.
\begin{proposition} \label{lemma:tracking_robustness_no_perturbation}
    Suppose $\gamma(\cdot)$, $r(\cdot)$ and $\hat{r}(\cdot)$ satisfy Proposition \ref{lemma:robustness_no_perturbation}. 
    If, for each $i \in \{1, ..., N\}$, the robot's coordinate under $\hat{r}(s_{i})$ matches exactly $\gamma(s_{i})$,
    then the trajectory planner is robust for tracking $\gamma(\cdot)$ while respecting dynamic feasibility.
\end{proposition}

Deriving a finite-time BRT for the quadrotor, 
an underactuated non-linear system,
is non-trivial. 
In this work, we approximate the BRT via a sampling-based approach.
Let $\psi(x_{0}, x, \Delta t)$ be a procedure that applies the low-level controller $U$ from initial state $x_{0}$ to verify whether it can reach target state $x$ within time $\Delta t$.
Formally, $\xi_{c}(x, \Delta t)$ comprises all $x_{0}$ for which $\psi(x_{0}, x, \Delta t)$ holds.
Hence: 
\begin{align}
    \mathbb{E}_{r(s_{i}), \hat{r}(s_{i})\sim\gamma} \Big[ \psi\Bigg(\hat{r}(s_{i}), r(s_{i+1}), t_{i})\Bigg) \Big] = Pr\Big(\hat{r}(s_{i}) \in \xi_{c}(r(s_{i+1}), t_{i})\Big). \label{eq:robustness1}
\end{align} 
We estimate this probability by sampling $r(s_{i})$ and $\hat{r}(s_{i})$ from the model predictions over $\gamma$ in data and simulating $U$ to determine whether $\hat{r}(s_{i})$ can reach $r(s_{i+1})$ within $t_{i}$.

\subsubsection{Robustness against Perturbations}

Although (\ref{eq:robustness1}) offers a means to empirically access model robustness, 
the result may be biased due to the large sequence space of $\gamma$.
To address this, we evaluate the robustness of tracking $\gamma(\cdot)$ under minor perturbations on the input paths, providing a more practical measure of the planner robustness.

Let $\epsilon$ denote the perturbation scale.
Define $\pi_{\epsilon}(\gamma)$ as the family of geometric paths that deviate from $\gamma(\cdot)$ by at most $\epsilon$ at each discrete step
while staying within the class of piece-wise polynomial paths. 
\begin{proposition} \label{lemma:robustness_with_perturbation}
    Suppose $\gamma(\cdot)$ is a geometric path.
    If, for each $\hat{\gamma} \in \pi_{\epsilon}(\gamma)$ and $i \in \{1, ..., N-1\}$,
    \begin{align}
        \hat{r}(s_{i}) \in \xi_{c}(r(s_{i+1}), t_{i}), \text{ where } t_{i} = 2\Delta s/(\sqrt{h(s_{i})}+ \sqrt{h(s_{i+1})}),
    \end{align}
    with $r(\cdot)$, $\hat{r}(\cdot)$ and $h(\cdot)$ all corresponding to $\hat{\gamma}$,
    then the trajectory planner is $\epsilon$-robust with respect to dynamic feasibility.
\end{proposition}
Correspondingly, we quantify $\epsilon$-robustness by sampling a set of perturbed geometric paths and simulating them under $U$ based on the predicted path parameterization.

\subsubsection{Robustness Enhancement via Noise Injection}

Proposition \ref{lemma:robustness_with_perturbation} naturally inspires a new training scheme that augments the dataset with randomized path perturbations. 
Rather than training exclusively on the original paths $\gamma(\cdot)$, 
we also include the perturbed paths $\hat{\gamma} \in \pi_{\epsilon}(\gamma)$ under a given perturbation scale $\epsilon$,
targeting to predict the same ground-truth $[h(\cdot), \cos\theta_{z}(\cdot)]$ at $\gamma$.
To ensure practical feasibility, we adopt the following assumption:
\begin{assumption}
    Let $\gamma(\cdot)$ be a geometric path, and $\epsilon$ be a perturbation scale. 
    For each $\hat{\gamma} \in \pi_{\epsilon}(\gamma)$, the control sequence $u(\cdot)$ that is optimal for $\gamma$ remains $\epsilon$-robust for $\hat{\gamma}$.
\end{assumption}
In essence, this assumption constrains how large $\epsilon$ can be.
Beyond model robustness, an excessively large $\epsilon$ would significantly compromise the time-optimal quality of the predictions.
While we do not derive a precise bound on $\epsilon$ in this work, 
we treat it as a tunable hyperparameter, 
and show the proposed training scheme's effectiveness in enhancing model robustness in Section \ref{sec:robustness_exp}.

\section{Experimental Results}
\label{sec:result}

\subsection{Simulation Experiments}

We construct the training dataset by generating minimum-snap trajectories through randomly sampled waypoints within a given range that are subsequently discretized via $100$ equally spaced points.
For each path, we apply TOPPQuad to obtain a dynamically feasible, time-optimal path parameterization under a $5 m/s$ speed limit,
resulting in a dataset of $10,000$ trajectories.
All simulation experiments are conducted in RotorPy~\cite{folk2023rotorpy} configured with CrazyFlie 2.0 parameters~\cite{8046794}.

\begin{table*}[t]
\centering
\small
\setlength{\tabcolsep}{4pt}
\renewcommand{\arraystretch}{1.2} 
\begin{tabular}{c c | cc cc cc cc}
\toprule
\multirow{2}{*}{} 
  & \multirow{2}{*}{\textbf{TOPPQuad}}
  & \multicolumn{2}{c}{\textbf{LSTM}}
  & \multicolumn{2}{c}{\textbf{Transformer}}
  & \multicolumn{2}{c}{\textbf{ETransformer}}
  & \multicolumn{2}{c}{\textbf{MLP}}\\
\cmidrule(lr){3-4}\cmidrule(lr){5-6} \cmidrule(lr){7-8} \cmidrule(lr){9-10}
  & 
  & \textbf{Train} & \textbf{Test} 
  & \textbf{Train} & \textbf{Test}
  & \textbf{Train} & \textbf{Test}
  & \textbf{Train} & \textbf{Test}\\
\midrule
\textbf{max dev (m)} & 0.053   & \textbf{0.074} & 0.143   & 0.607 & 0.649 & 0.195 & 0.226 & 0.252 & 0.305 \\
\textbf{thrust violation (N)} & 0.000 & \textbf{0.002} & 0.009 & 0.135 & 0.123 & 0.012 & 0.018 & 0.031 & 0.048 \\
\midrule
\textbf{TD ratio (\%)} & 5.929(s) & -0.70\% & \textbf{-0.40\%} & -8.50\% & -2.35\% & 1.89\% & 2.49\% & -6.59\% & -3.03\% \\
\midrule
\textbf{failure (\%)} & 0.0\% & 2.0\% & 4.0\% & 76.0\% & 72.0\% & 6.0\% & 4.0\% & \textbf{0.0\%} & 6.0\% \\
\textbf{compute time (s)} & 10.656 & 0.078 & 0.096 & 1.012 & 1.042 & 0.010 & 0.018 & \textbf{0.005} & 0.014 \\
\bottomrule
\end{tabular}
\caption{\small 
    \textbf{Ablation study on model architectures.}
    Each statistic is averaged over $100$ trajectories.
    A negative \textit{TD ratio}
    ($\frac{\text{PRED. TIME} - \text{OPT. TIME}}{\text{OPT. TIME}}$) denotes shorter travel time relative to TOPPQuad (OPT. TIME). 
    Although the low-level controller enforces per-motor limits,
    the nonlinear mapping from the inputs of the  TOPP problem to predicted motor thrusts is not bounded by construction. 
    Hence, the approximate nature of learning-based methods can lead to faster path execution times, albeit with 
    non-zero thrust violations. 
    A key desideratum of such methods is to minimize thrust violations without producing overly conservative (slow) trajectories. 
}
\label{table:ablation}
\vspace{-10pt}
\end{table*}

\subsubsection{Architecture Ablation}

We begin by describing the candidate architectures.
\textbf{LSTM Encoder-Decoder}~\cite{lstm,lstm-enc-dec} uses an LSTM encoder (with a  non-parameterized attention mechanism equipped) mapping the input trajectory to a latent representation, 
and an LSTM decoder to generate outputs auto-regressively.
\textbf{Transformer Encoder-Decoder} (denoted as \textit{Transformer})~\cite{transformer} uses self-attention in the encoder to capture intra-sequence dependencies, 
and cross-attention in the decoder,
trained via teacher forcing~\cite{bengio2015scheduled}, i.e., masking the shifted ground-truth output as the decoder input.
\textbf{Encoder-Only Transformer} (denoted as \textit{ETransformer}) removes the decoder altogether
to obviate the need for teacher forcing.
Finally, \textbf{Per-Step MLP} is a multilayer perceptron that predicts the outputs at each discrete step individually.
Detailed implementation details and additional ablation studies appear in the Appendix.

To evaluate the trajectory planners, 
we measure the \textit{maximum deviation} in position from the reference trajectories,
average \textit{thrust violation} indicating adherence to actuation constraints,
and time-optimality by comparing travel Time Difference ratio (\textit{TD ratio}) with respect to TOPPQuad.
An attempt is classified as a \textit{failure}
if it leads to a crash, 
or if the maximum position deviation exceeds $1$ meter. 
We also report the \textit{compute time} required by each planner.

Table \ref{table:ablation} presents the ablation results.
\textbf{LSTM} achieves the best performance among all candidates,
with a maximum position deviation only $0.023$m above TOPPQuad and negligible thrust violation,
resulting in an almost zero failure rate.
The slight reduction in travel time arises because the LSTM occasionally yields velocities marginally above the $5 m/s$ speed limit.
In contrast, \textbf{Transformer} has worse tracking accuracy and higher failure rates, 
consistent with known difficulties in training transformers via teacher forcing with limited data \cite{bengio2015scheduled}. This aligns with the larger TDRatio, where a faster travel time necessitates greater thrust bound violations.
\textbf{ETransformer} provides competitive results but still lags behind the LSTM in tracking accuracy and time optimality. 
Finally, \textbf{Per-Step MLP} struggles to track the reference trajectory precisely and incurs high thrust violations
as it must repeatedly base its predictions on its own prior outputs, 
leading to out-of-distribution issues.

\begin{table*}[t]
\centering
\small
\setlength{\tabcolsep}{4.5pt} 
\renewcommand{\arraystretch}{1.2} 
\begin{tabular}{c ccc ccc ccc}
\toprule
\multirow{2}{*}{}  
  & \multicolumn{3}{c}{\textbf{LSTM}}
  & \multicolumn{3}{c}{\textbf{LSTM-0.01}}
  & \multicolumn{3}{c}{\textbf{LSTM-0.1}} \\
\cmidrule(lr){2-4}\cmidrule(lr){5-7}\cmidrule(lr){8-10} 
$\epsilon$ (perturbation scale) & 0.001 & 0.01 & 0.1  & 0.001 & 0.01 & 0.1 & 0.001 & 0.01 & 0.1 \\
\midrule
\textbf{max deviation (m)} & 0.234 & 0.790 & 0.739 & 0.094 & \textbf{0.093} & 0.448 & 0.127 & 0.126 & 0.127 \\ 
\textbf{TD ratio (\%)} & -0.13\% & 1.33\% & 3.21\% & 0.08\% & 0.09\% & 3.73\% & -0.07\% & \textbf{0.04\%} & 0.29\% \\ 
\textbf{output variation} & 0.005 & 0.206 & 0.306 & \textbf{0.000} & 0.005 & 0.089 & 0.001 & 0.002 & 0.013 \\ 
\textbf{in-BRT probability (\%)} & 90.0\% & 78.8\% & 70.0\% & 94.3\% & \textbf{94.5\%} & 91.4\% & 92.4\% & 93.0\% & 92.9\% \\ 
\bottomrule
\end{tabular}
\caption{\small 
    \textbf{Robustness analysis for the LSTM Encoder-Decoder model. }
    Each statistic is averaged over $100$ trajectories with $10$ sets of perturbations randomly drawn per trajectory.
    \textit{TD ratios} are computed relative to TOPPQuad's optimal solutions (not shown in the table).
    Non-percentage values are rounded to three decimal places, so `$0.000$' does not imply an exact zero.
}
\label{table:robustness}
\vspace{-10pt}
\end{table*}

\subsubsection{Robustness Analysis}
\label{sec:robustness_exp}

Next, we conduct a robustness analysis on the LSTM encoder-decoder model. 
We evaluate both the model trained solely on clean data
and two variants, \textit{LSTM-0.01} and \textit{LSTM-0.1},
that incorporate randomized path perturbations of scales $0.01$ and $0.1$, respectively, as to augment training data.
To evaluate robustness, 
we apply controlled perturbations to the input geometric paths. 
Two additional metrics are reported. 
First, \textit{output variation} quantifies the model sensitivity,
computing the average of the maximum absolute differences in model outputs. 
Second, \textit{in-BRT probability} assesses dynamic feasibility of the predicted path parameterization, as defined in Section \ref{sec:robustness_analysis}

Table \ref{table:robustness} presents the robustness analysis results.
When trained exclusively on clean data,
the model is highly sensitive to input perturbations. 
This is reflected by the large values in output variation.
Even under small input perturbations such as $\epsilon=0.001$, its maximal position deviation rises from $0.143$ m (Table \ref{table:ablation}) to $0.234$ m.
Larger perturbations further degrade tracking performance
and reduce in-BRT probability, which indicates a greater likelihood of generating dynamically infeasible predictions.

In contrast, the models trained with augmented noisy data yields improved output stability and in-BRT probability.
Notably, LSTM-0.1 consistently maintains low maximum position deviations and high in-BRT probabilities at all tested perturbation levels,
highlighting its strong robustness.
LSTM-0.01 also shows enhanced robustness up to its training perturbation level ($\epsilon=0.01$).
Note that a higher in-BRT probability does not imply superior tracking,
as it primarily measures the dynamic feasibility of the predicted trajectory under quadrotor dynamics.
Furthermore, LSTM-0.1 trades off some time-optimality and tracking accuracy for robustness, evident when examining low-perturbation regimes ($\epsilon=0.001$).
Training over a broader neighborhood of nominal paths ensures robust feasibility but can shift the solution away from the exact time-optimal trajectories. 
This suggests 
that an upper limit to training perturbation levels must exist in practice.

\subsubsection{Baseline Comparison}

We compare the performance of the proposed algorithm against two learning-based baselines. 
Time Allocation Network (\textbf{AllocNet})~\cite{wu2024deep} processes a point cloud of obstacles to predict time-optimal trajectories for safe navigation between specified start and goal locations. 
Obstacles are constructed from a sequence of bounding boxes placed around the reference trajectory.
Multi-Fidelity Black-Box Optimization Trajectory Planner (\textbf{MFBOTrajectory})~\cite{ryou2021multi} employs Gaussian Processes to allocate time between waypoints, 
requiring iterative online simulations to handle unseen waypoint arrangements.
These methods aim to generate time-optimal trajectories within the class of polynomials, so the resulting paths may deviate significantly from the TOPPQuad and LSTM path. Examples are provided in the Appendix.
Hence, the quadrotor’s \textit{average speed} and mean \textit{path length} are reported.

Table \ref{table:baseline} summarizes this comparison. 
AllocNet suffers from a higher failure rate due to a mismatch between predicted time allocations and the polytopes used in the safe flight corridor during prediction.
Notably, our testing setup uses generous bounding boxes to avoid distribution mismatch issues.
Among its successful runs,
AllocNet's predicted trajectories achieve shorter travel times by finding shorter geometric paths, yet yields a lower average speed than the LSTM.
MFBOTrajectory achieves zero failures,
but produces the slowest average speed,
leading to longer travel time despite formulating shorter polynomial paths than the LSTM.
Moreover, MFBOTrajectory must retrain online with each new unseen waypoint arrangement,
incurring significant computational overhead.
Unlike the LSTM, neither approach utilizes the full flight profile of the quadrotor, as seen by the commanded minimum and maximum thrusts.

\begin{table*}[t]
\centering
\small
\setlength{\tabcolsep}{6pt} 
\renewcommand{\arraystretch}{1.2} 
\begin{tabular}{c c | c c c c}
\toprule
  & \textbf{TOPPQuad}
  & \textbf{LSTM} 
  & \textbf{LSTM-0.1}
  & \textbf{AllocNet}
  & \textbf{MFBOTrajectory} \\ 
\midrule
\textbf{max deviation (m)} & 0.039 & 0.124 & 0.170 & 0.037 & \textbf{0.010} \\
\textbf{min/max thrusts (N)} & 0.05 / 0.14 & 0.03 / 0.16 & 0.03 / 0.17 & 0.06 / 0.10 & 0.07 / 0.08 \\ 
\textbf{thrust violation (N)} & 0.000 & 0.002 & 0.002 & \textbf{0.000} & \textbf{0.000} \\
\midrule
\textbf{traj time (s)} & 5.687 & 5.653 & 5.691 & \textbf{4.066} & 9.431 \\
\textbf{path length (m)} & 19.772 & 19.772 & 19.772 & \textbf{11.453} & 16.778 \\
\textbf{average speed (m/s)} & 3.477 & \textbf{3.498} & 3.474 & 2.817 & 1.779 \\
\midrule
\textbf{failure rate (\%)} & 0\% & \textbf{0\%} & \textbf{0\%} & 28\% & \textbf{0\%} \\
\textbf{compute time (s)} & 10.083 & \textbf{0.069} & 0.101 & 0.277 & 13609.913 \\
\bottomrule
\end{tabular}
\caption{\small 
    \textbf{Baseline comparison. }
    Each statistic is averaged over 50 trajectories.
    Compute time is measured from when the target path (or waypoints) are provided until a valid path parameterization is returned.
    For \textbf{MFBOTrajectory}, which undergoes online retraining for each unseen waypoint configuration, 
    the retraining time is included for a fair comparison.
}
\label{table:baseline}
\vspace{-10pt}
\end{table*}

\subsection{Hardware Experiments}

Next, we validate our approach on hardware using a CrazyFlie 2.0 quadrotor tracked in a Vicon motion capture space.
For safety,
we limit the maximum speed to $2$ $m/s$, the maximum acceleration to $10$ $m/s^2$, and the maximum angular velocity to $10$ $rad/s$.
Under these parameters,
we generate a new dataset of $9,000$ trajectories in simulation,
following the same procedures used in our earlier experiments.
We train an LSTM encoder-decoder model with a perturbation scale of $0.001$.
Experiments with other training configurations are provided in Appendix.

\begin{figure}[h]
    \centering
    \includegraphics[width=1\linewidth]{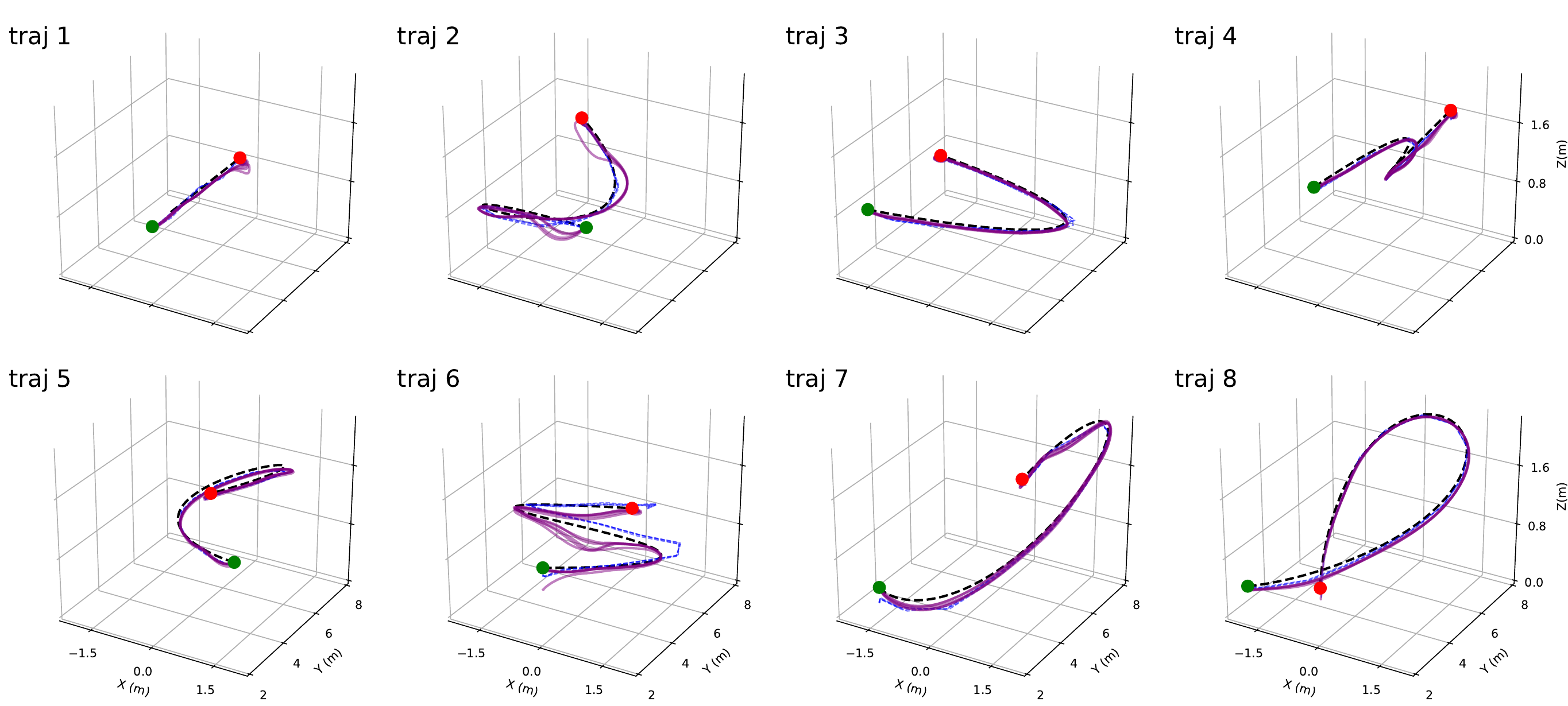}
    \caption{
        \small Experiment visualization, plotted from the collected motion capture flight data. 
        The dashed black line represents the reference geometric path, the dashed blue line shows the tracked trajectory generated by TOPPQuad, and the solid purple lines show the tracked trajectories output by our model across different runs.
    }
    \label{fig:hardware}
\vspace{-10pt}
\end{figure}

Table \ref{table:hardware} shows quantitative statistics, while Figure \ref{fig:hardware} provides visualizations.
We conduct tests on eight distinct geometric paths, 
with four trials of TOPPQuad and five trials of our proposed method.
As in simulation, our method yields position deviations comparable to TOPPQuad.
However, the travel time difference ratio increases.
This discrepancy arises due to the difficulty of settling the quadrotor at its goal, 
an issue exacerbated in the learning-based approach by the sim-to-real gap.

\begin{figure}[!tbp]
  \centering
  \subfloat[]{\includegraphics[width=0.4\textwidth]{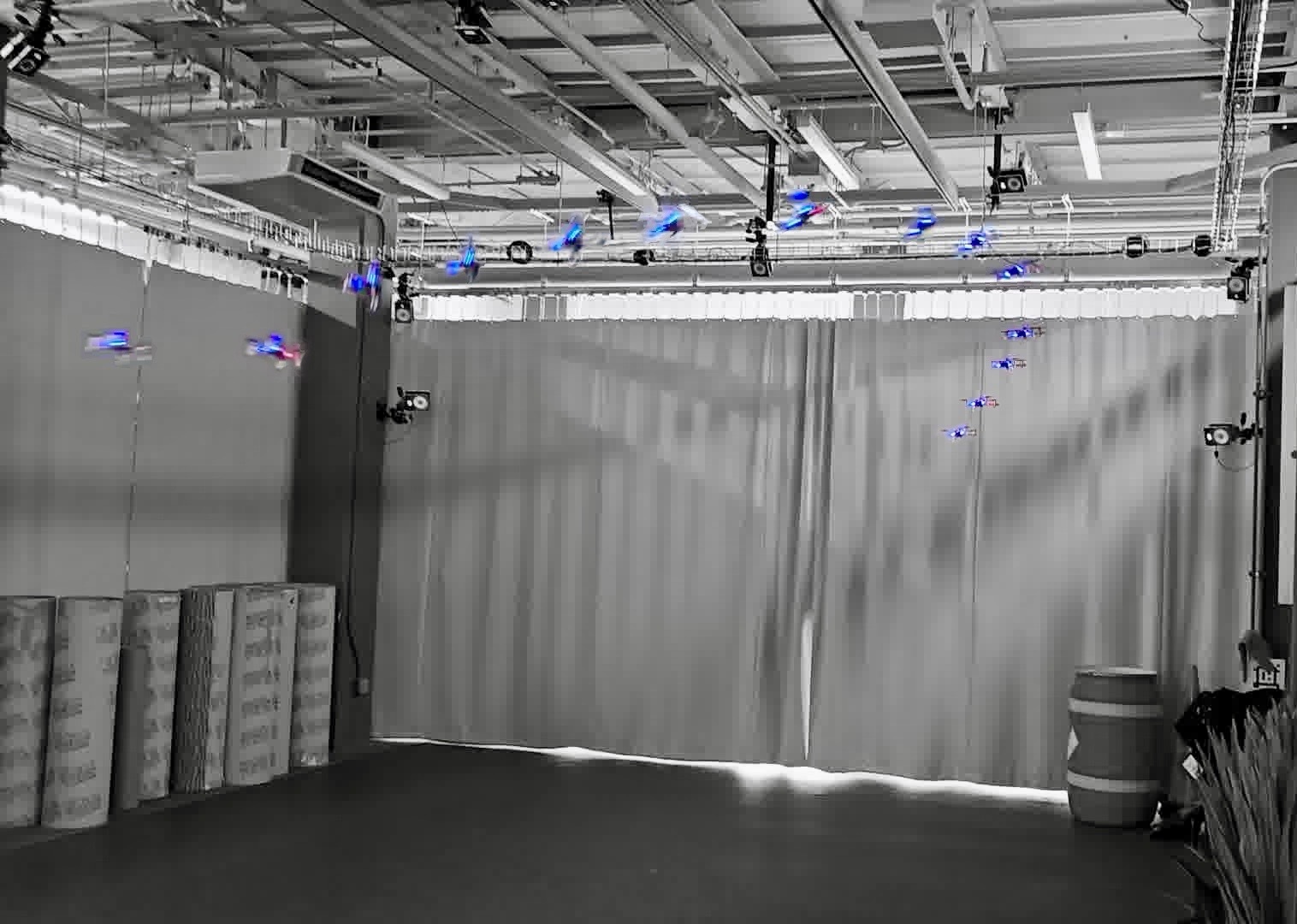}\label{fig:f1}}
  \hfill
  \subfloat[]{\includegraphics[width=0.6\linewidth]{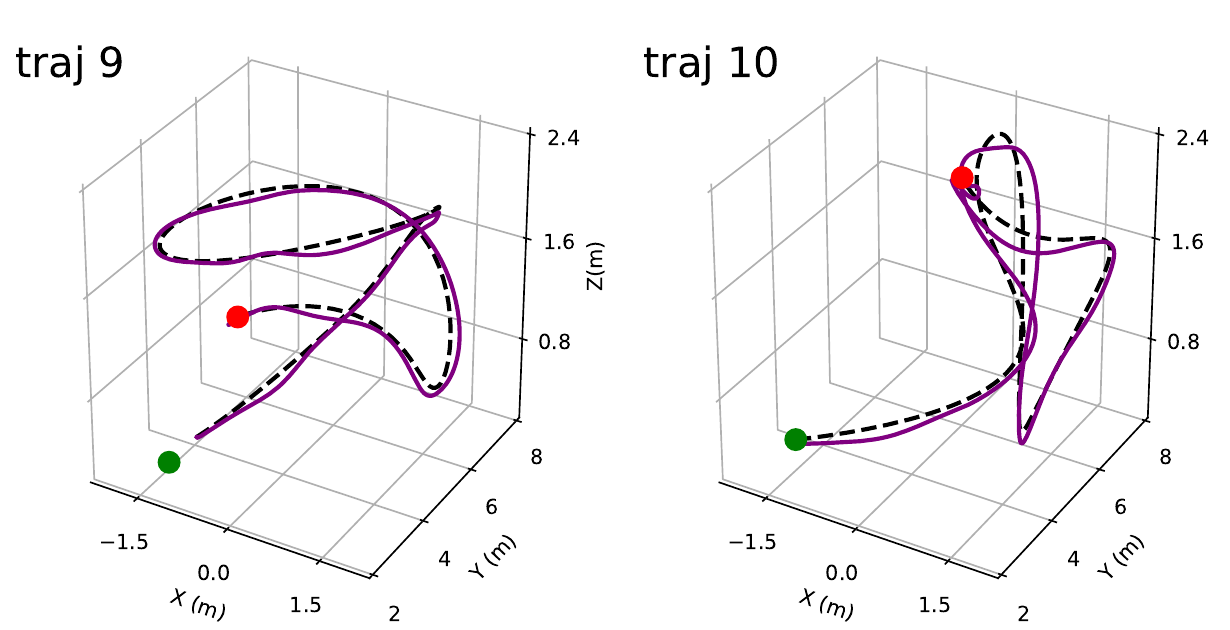}}
  \caption{\small
        a) Partial depiction of Traj 9. b) Experiment Visualization, 
        showcasing good tracking of geometric paths whose lengths exceede what is seen in the dataset
    }
 \label{fig:hardware_long_sequence}
\end{figure}

Moreover, the proposed method naturally generalizes to longer sequences than those encountered in training.
The key modification is to have the model  predict only a segment of the overall trajectory at once, 
while conditioning on additional information capturing the robot's state at the start of each segment.
As sequence lengths grow, the compute speedups become increasingly substantial.
Figure \ref{fig:hardware_long_sequence} demonstrates the resulting extended trajectory predictions.

\begin{table*}[t]
\centering
\small
\setlength{\tabcolsep}{4.5pt} 
\renewcommand{\arraystretch}{1.2} 
\begin{tabular}{c c c}
\toprule
 & \textbf{TOPPQuad} & \textbf{Learning-based TOPPQuad } \\
\midrule
\textbf{max deviation (m)} & 0.347 & 0.355  \\ 
\textbf{travel time (s)} & 7.981 & 8.355 \\  
\bottomrule
\end{tabular}
\caption{\small Hardware experiments.
}
\label{table:hardware}
\vspace{-10pt}
\end{table*}

\section{Conclusion}
\label{sec:conclusion}

In this work, we propose an imitation-learning framework for time-optimal quadrotor trajectory generation.
Guided by domain-specific insights, we introduce a concise yet effective input-output feature design.
We also present a rigorous robustness analysis framework alongside a data augmentation strategy that enhances model robustness.
Through comprehensive studies, our method is shown to closely imitate a model-based trajectory planner, 
producing near-optimal solutions that largely respect dynamic feasibility and delivering a significant computational speedup over similar optimization-based methods. 
Finally, we validate the approach on a hardware quadrotor platform, demonstrating its practical effectiveness.

\subsection{Limitations}

A key limitation of the proposed approach is that the model must be retrained whenever the quadrotor's configuration changes, as its learned parameters are tailored to specific hardware setups and dynamic constraints from the training dataset.
Furthermore, while the model generally approximates the expert solution accurately, the approximate nature of learning-based methods may still yield infeasible path parameterizations.
Finally, the optimal perturbation bounds for data augmentation remain an open question.

\bibliography{references}  

\end{document}


\appendix
\section*{Appendix}

\section{Dataset Generation}

\subsection{TOPPQuad Modifications and Dataset Generation}
\label{sec:dataset}

\begin{figure}[h]
    \centering
    \includegraphics[width=\linewidth]{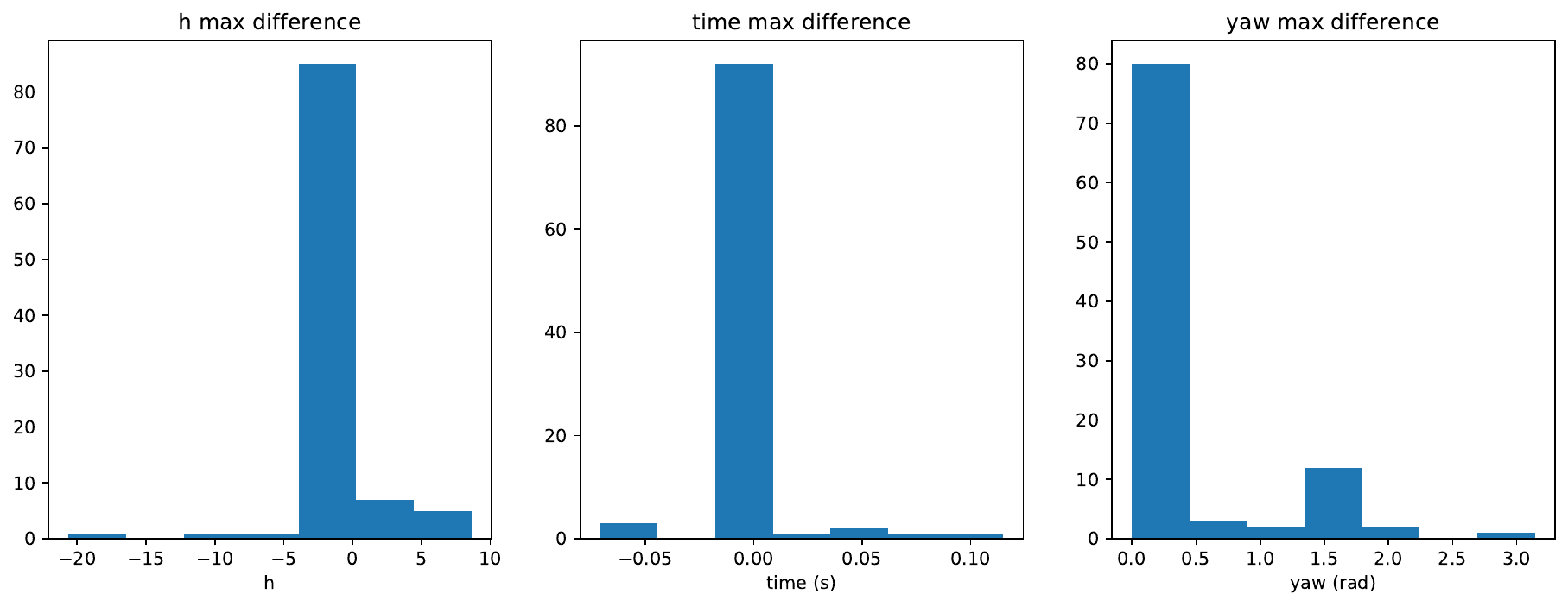}
    \includegraphics[width=\linewidth]{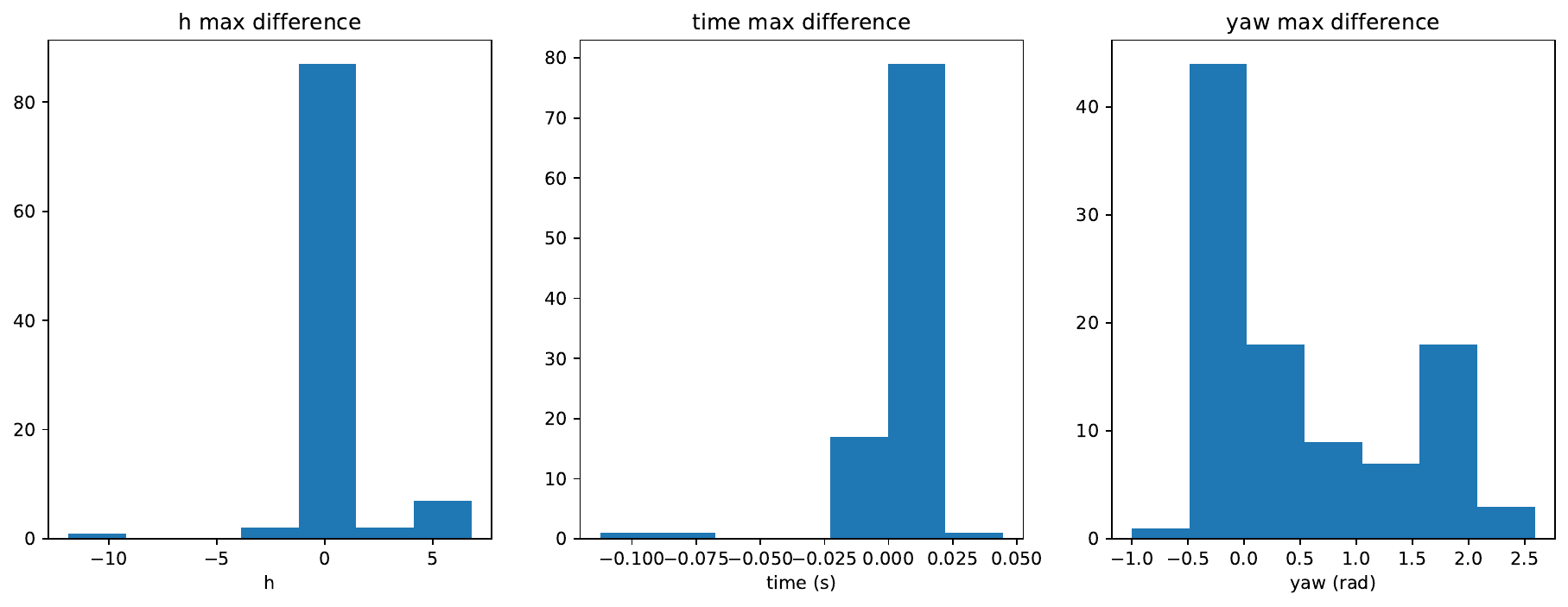}
    \caption{Maximum difference of h, time, and yaw between 100 pairs of perturbed and unperturbed TOPPQuad trajectories. \textbf{Top}: TOPPQuad trajectory comparison computed following \cite{10801611}. \textbf{Bottom}: TOPPQuad trajectory comparison computed with the modifications introduced in Sec \ref{sec:dataset}. Both rows are computed with the $v_{max} = 5m/s$ constraint.}
    \label{fig:yaw_consistency}
\end{figure}

The LSTM models are trained on a dataset of trajectories generated from TOPPQuad \cite{10801611}. On top of the dynamic feasibility constraints described in the paper (most notably bounds on the individual motor thrusts), we enforce a maximum velocity constraint $v_{max}$ and constrain the starting yaw angle $\theta_{z,0} \in [0, \pi/2]$ to account for symmetries around the body z-axis of the quadrotor. 
Additionally, we add a small penalty on the total angular acceleration to the cost function to ensure consistency of optimal yaw trajectories across geometric paths that only differ by small perturbations. 
Therefore, trajectories in the dataset minimize the following objective function:

\begin{align}
    \min \
     T + \lambda \sum_{i = 0}^{N-1} ||\alpha_{i}||_{2}^{2},
\end{align}
where
\begin{align}
    T = \sum_{i = 0}^{N-1} \frac{2\Delta s}{\sqrt{h_i} + \sqrt{h_{i+1}}}
\end{align}
and $N$ is the number of path discretization points. We select $\lambda = 10^{-4}$. 
In Fig. \ref{fig:yaw_consistency}, we compare the maximum difference between yaw profiles of TOPPQuad trajectories with and without this added penalty. For each datapoint, we use TOPPQuad to compute a time-optimal trajectory with and without a $[0.1, 0, 0]m$ perturbation to the first waypoint. Despite a minimal change in the h profile and total trajectory time, the maximum yaw difference clusters much more tightly around 0 with this added penalty. We posit this is due to maximum velocity being the dominant constraint in the optimization, preventing motor saturation for stretches along the path. This in turn provides the thrust authority for additional yawing without significantly impacting the total traversal time along the path. Although such an effect is innocuous when optimizing for a single trajectory, the effect is undesirable for the learning problem.

\subsubsection{Simulation Experiment Dataset}

For the simulation experiments, we set a maximum velocity $v_{max} = 5m/s$ for a fair comparison to the selected baselines. Waypoints are sampled from a $10m \times 10m \times 10m$ box. The models are trained on a dataset of approximatley $10,000$ trajectories. The trajectories for the baseline comparisons have waypoints sampled from a $10m \times 10m \times 5m$ box, again to ensure a fair comparison across baselines.

\subsubsection{Hardware Experiment Dataset}

For the hardware expeirments, we set a maximum velocity of $v_{max} = 2m/s$, a maximum angular velocity of $\omega_{max} = 8 rad/s$, and a maximum acceleration of $a_{max} = 5m/s^2$. These numbers were selected based on observation of what the CrazyFlie was physically capable of achieving, so as to minimize model mismatch. Again, waypoints for the dataset were sampled from a $10m \times 10m \times 10m$ box. Trajectories for the experiments have waypoints sampled from a $4.5m \times 8m \times 2m$ box to fit the Vicon motion capture space.

\section{Trajectory Recovery}

Given the path of positions ($\gamma(\cdot)$), together with its first ($\gamma'(\cdot)$) and second ($\gamma''(\cdot)$) derivatives, our model outputs the square speed profile ($h(\cdot)$) and a cosine-normalized yaw ($cos \theta_z(\cdot)$). Here, we show a sketch of how to recover the full trajectory variables from these outputs for the geometric controller \cite{se3control}. We refer the reader to the \cite{10801611} for full details.

\textbf{Translational Components}: Given the problem formulation, we can immediately state $\mathbf{p}(t) = \gamma(\cdot)$. Differentiating this equation, we  get velocity and acceleration as
\begin{align}
    \label{eq:vel_scale}
    \mathbf{v}(t) = \sqrt{h(\cdot)}\gamma'(\cdot), \\
    \label{eq:acc_scale}
    \mathbf{a}(t) = \frac{1}{2} \gamma'(\cdot) h'(\cdot) + \gamma''(\cdot) h(\cdot).
\end{align}

\textbf{Rotational Components}: We represent the orientation of the quadrotor using the Hopf Fibration \cite{watterson2019control}, where the rotation quaternion can be broken down into quaternions representing the quadrotor's body-z axis ($\mathbf{b}_3$) and the yaw angle. 
\begin{align}
    \mathbf{q} = \mathbf{q}_{b_3} \otimes \mathbf{q}_{yaw}
\end{align}

Starting from the quadrotor's equations of motion, we know the thrust vector ($\mathbf{F}$) is aligned with ($\mathbf{b}_3$),
\begin{align}
    \mathbf{F} = m\mathbf{a} - \mathbf{g}
\end{align}
where $m$ the mass of the quadrotor and $\mathbf{g}$ the gravity vector. Taking the unit vector of $\mathbf{F}$ gets us $\mathbf{b}_3$, which can be computed from the previously determined acceleration (Eq. \ref{eq:acc_scale}).
\begin{align}
    \mathbf{b}_3 = \frac{\mathbf{F}}{||\mathbf{F}||} = \frac{ m\mathbf{a} - \mathbf{g}}{|| m\mathbf{a} - \mathbf{g}||} = \begin{pmatrix} a \\ b \\ c \end{pmatrix}
\end{align}
Using the Hopf Fibration, we re-write this into quaternion form as 
\begin{align}
    \mathbf{q}_{b_3} = \frac{1}{\sqrt{(2(1+c)}}\begin{bmatrix} 1+c & -b & a & 0 \end{bmatrix}^T
\end{align}
The yaw quaternion is written as
\begin{align}
    \mathbf{q}_{yaw} = \begin{bmatrix} cos(\theta_z/2) & 0 & 0 & sin(\theta_z/2) \end{bmatrix}^T
\end{align}
which can be directly populated from the model output. We note that by predicting only $cos \theta_z$, we lose information on the direction of yaw change at the $\theta_z = \pi$ boundary. However, we make the assumption that because of the added penalty on $\alpha_z$ in the cost function during dataset generation (Sec \ref{sec:dataset}), any change in yaw will be minimized and the yaw profile will not make sudden sharp jumps. We compute the yaw trajectory accordingly.
Finally, we compute the angular velocity ($\omega$) by inverting Eq. 23 from \cite{10801611}, rewritten here:
\begin{equation} \label{eq:quaternion_discrete_dynamics}
\mathbf{q}_{i+1} = \frac{(\mathbf{I}_4 + \frac{\Delta s}{2} \Omega(\omega_i)) \ \mathbf{q}_i}{\sqrt{1 + \frac{\Delta s^2}{4} || \omega_i ||_2^2}},
\end{equation}
where $\Delta s$ is the spacing between consecutive discretized points. By recognizing $\Omega(\omega_i)$ is the matrix multiplication equivalent to a quaternion multiplication for rotation velocity \cite{graf2008quaternionsdynamics} (and likewise treating $\mathbf{I}_4$ as an identity quaternion), we can multiply $\mathbf{q}_i$ into the numerator and set the denominator to some arbitrary variable $f$ 
\begin{align}
    \mathbf{q}_{i+1} = \frac{1}{f} \left( \mathbf{q}_i + \frac{\Delta s}{2}\mathbf{q}_i \otimes \begin{bmatrix} 0 \\ \omega_i  \end{bmatrix} \right)  = 
    \frac{1}{f}
    \mathbf{q}_i \otimes
    \begin{bmatrix}
        1 \\ 
        \frac{\Delta s }{2}\omega_{i}
    \end{bmatrix}
    .
\end{align}

Pre-multiplying both sides by $\mathbf{q}_i^{-1}$
\begin{align}
    \mathbf{q}_i^{-1}\otimes \mathbf{q}_{i+1} = 
    \frac{1}{f} \ \mathbf{q}_i^{-1} \otimes \mathbf{q}_i \otimes
    \begin{bmatrix}
        1 \\ 
        \frac{\Delta s }{2}\omega_{i}
    \end{bmatrix} = 
    \frac{1}{f}
    \begin{bmatrix} 
    1 \\ 
    \frac{\Delta s \ \omega_i}{2}  
    \end{bmatrix}.
    \label{eq:almost_invert}
\end{align}

Let $\mathbf{q}_i^{-1}\otimes \mathbf{q}_{i+1} = \begin{bmatrix} q_w \\ \mathbf{q}_v  \end{bmatrix}$. We use the scalar component of this quaternion to solve for $f$ then isolate $\omega_i$ from Eq.\ref{eq:almost_invert} to get
\begin{align}
    \omega_i(\cdot) = \frac{2\mathbf{q}_v}{\Delta s \ q_w}.
\end{align}
Re-scaling $\omega_i(\cdot)$ with $h$ to fit the learned time profile as a parallel to Eq. \ref{eq:vel_scale}, we finally get
\begin{align}
    \omega_i(t) = \sqrt{h}\frac{2\mathbf{q}_v}{\Delta s \ q_w}.
\end{align}

\newpage

\section{Additional Experiments}

\subsection{Input Ablation Study}

The goal of this experiment is to compare the performance of the LSTM architecture given different sets of inputs. Using the same metrics as prior ablation experiments, we compare the performance of LSTM models trained on combinations of the position, initial velocity, and initial acceleration of dataset trajectories in Table \ref{table:input-ablation}. By all four metrics, its clear that position alone is insufficient. The combinations of position, velocity, and acceleration all perform similarly, but we select the \textbf{pos/vel/acc} model due to its lower TD Ratio and failure rate with minimal losses in simulation performance. 

\begin{table*}[h]
\centering
\small
\setlength{\tabcolsep}{4pt} 
\renewcommand{\arraystretch}{1.2} 
\begin{tabular}{c c | cc cc cc cc}
\toprule
\multirow{2}{*}{} 
  & \multirow{2}{*}{\textbf{TOPPQuad}}
  & \multicolumn{2}{c}{\textbf{pos/vel/acc}}
  & \multicolumn{2}{c}{\textbf{pos/vel}}
  & \multicolumn{2}{c}{\textbf{pos/acc}}
  & \multicolumn{2}{c}{\textbf{pos}}\\
\cmidrule(lr){3-4}\cmidrule(lr){5-6} \cmidrule(lr){7-8} \cmidrule(lr){9-10}
  & 
  & \textbf{Train} & \textbf{Test} 
  & \textbf{Train} & \textbf{Test}
  & \textbf{Train} & \textbf{Test}
  & \textbf{Train} & \textbf{Test}\\
\midrule
\textbf{max dev (m)} & 0.053   & 0.074 & 0.143   & 0.106 & 0.148 & \textbf{0.068} & 0.164 & 0.097 & 0.225 \\
\textbf{thrust violation (N)} & 0.000 & \textbf{0.002} & 0.009 & \textbf{0.002} & 0.005 & 0.003 & 0.007 & 0.008 & 0.026 \\
\midrule
\textbf{TD ratio (\%)} & 5.929(s) & -0.70\% & \textbf{-0.40\%} & -0.70\% & -1.9\% & -0.90\% & -2.10\% & -0.70\% & -1.90\% \\
\midrule
\textbf{failure (\%)} & 0.0\% & \textbf{2.0\%} & 4.0\% & 4.0\% & 4.0\% & \textbf{2.0\%} & 6.0\% & 8.0\% & 22.0\% \\
\bottomrule
\end{tabular}
\caption{\small 
    \textbf{Ablation study on model architectures.}
    Each statistic is averaged over $100$ trajectories.
    A negative \textit{TD ratio}
    ($\frac{\text{PRED. TIME} - \text{OPT. TIME}}{\text{OPT. TIME}}$) denotes a shorter travel time relative to TOPPQuad (OPT. TIME). 
}
\label{table:input-ablation}
\vspace{-10pt}
\end{table*}

\subsection{Baseline Paths}

\begin{figure}[H]
    \centering
    \includegraphics[width=\linewidth]{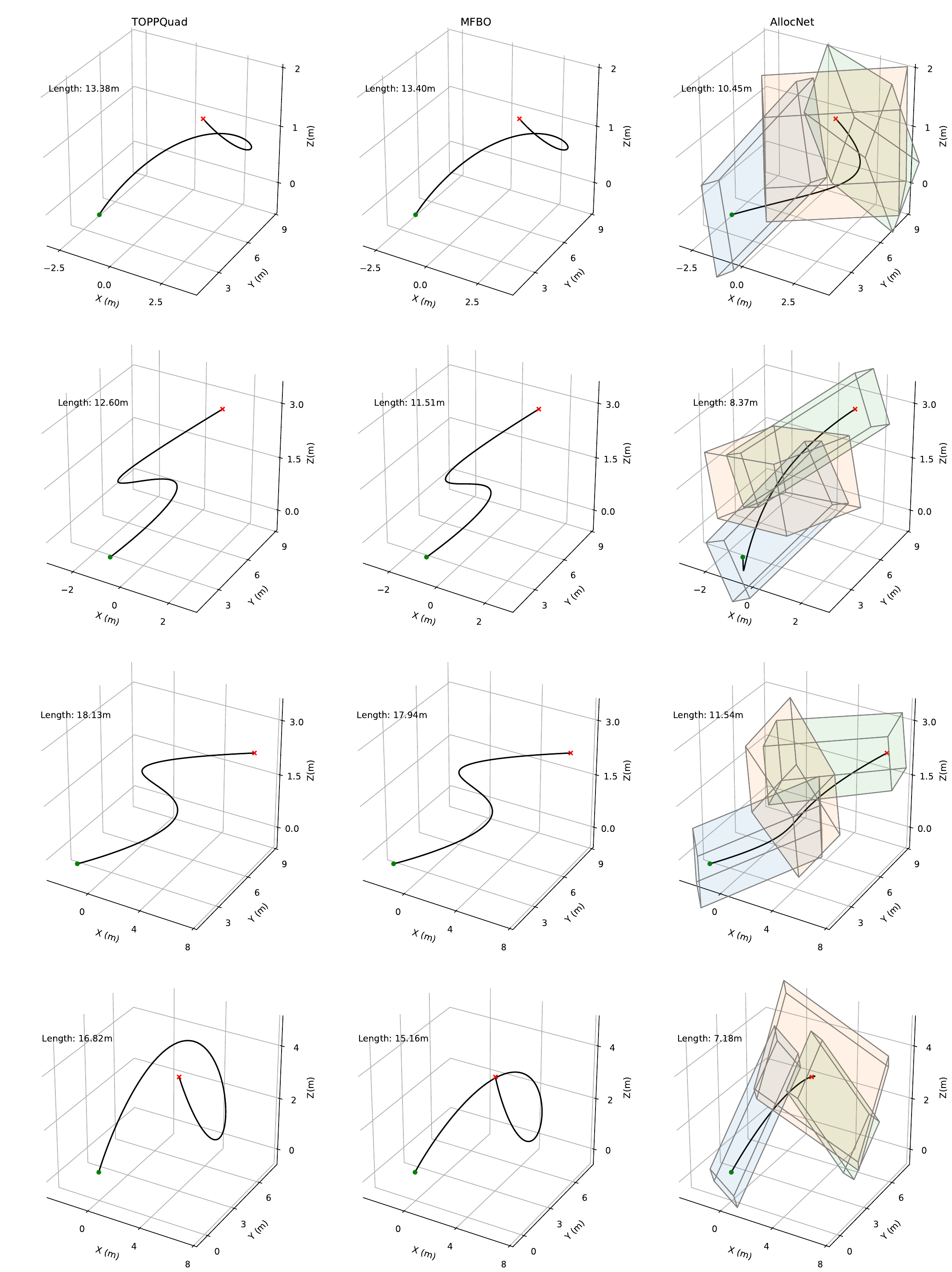}
    \caption{Sample paths generated by our planner (\textbf{left}), MFBO (\textbf{middle}) and AllocNet (\textbf{right}). We also show the bounding boxes given to the AllocNet planner.}
    \label{fig:baseline-paths}
\end{figure}

In this section, we select successful trajectories from all three baselines to illustrate the change in path shape and length (Fig. \ref{fig:baseline-paths}). Our planner (and TOPPPQuad) determines a dynamically-informed path using the minimum snap trajectory planner \cite{5980409}, given a set of waypoints and a set of time segments calculated by the euclidean distance between consecutive waypoints and a nominal velocity ($v_{nom} = 1m/s$). MFBO \cite{ryou2021multi} likewise uses a minimum snap planner with the same waypoints, but attempts to find a better set of time segments that is time optimal while respecting quadrotor dynamics. Finally, AllocNet \cite{wu2024deep} keeps the start and end waypoints for a minimum jerk trajectory, but attempts to predict both intermediary waypoints and the time allocation for a time-optimal trajectory from a set of convex corridors. 

For the AllocNet planner, we fit three bounding boxes to evenly spaced sections of the original minimum snap path, oriented to the plane of each path sections and with a $0.75m$ padding from any box wall to the path. The exterior of these bounding boxes are then converted into a points cloud, from which AllocNet generates a set of convex corridors. We select the padding to ensure the convex corridors generated are generous enough to not lie outside of the planner's training set for a fair comparison.

\newpage

\subsection{Noised Model Hardware Experiments}
\begin{figure}[h]
    \centering
    \includegraphics[width=0.95\linewidth]{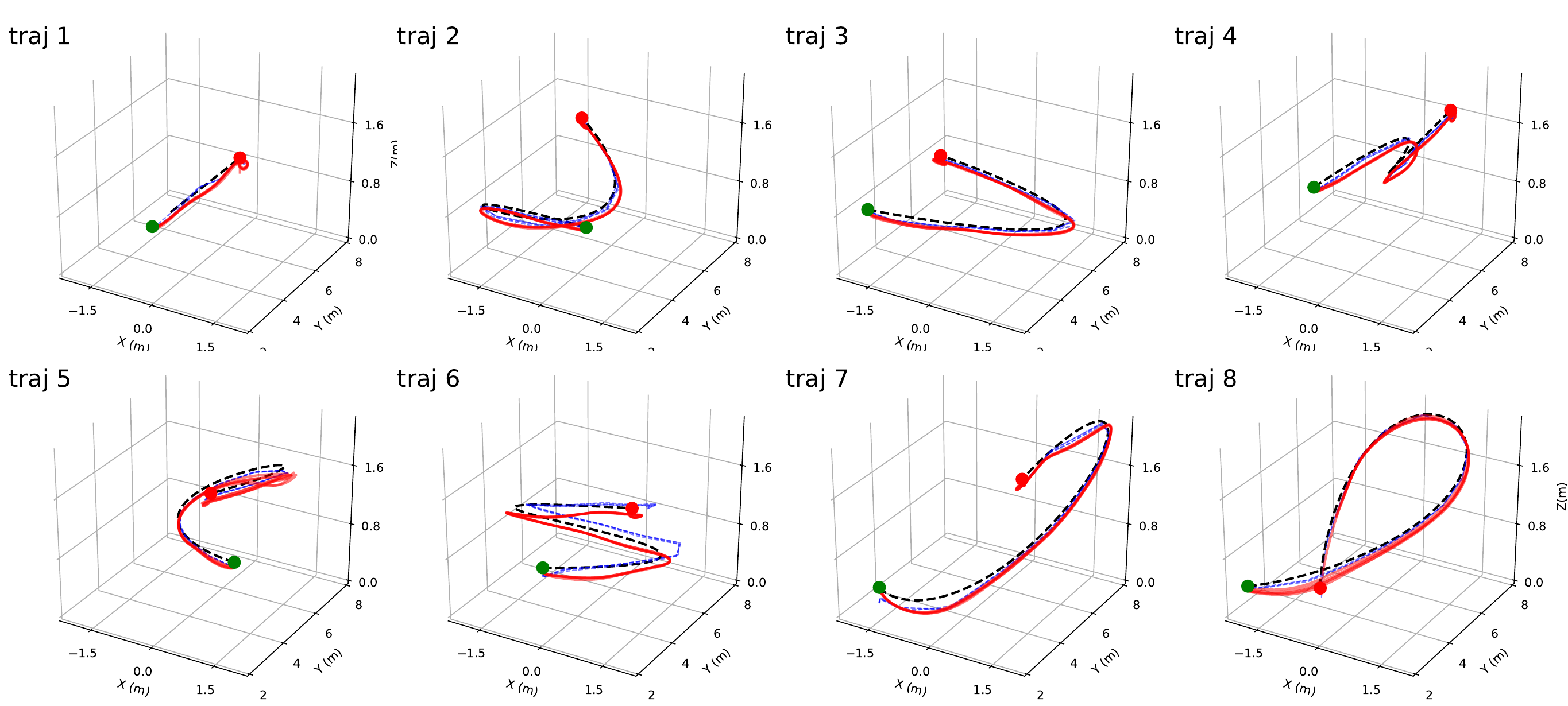}
    \caption{
        \small Experiment visualization of trajectories predicted from the \textbf{LSTM-0.1} model, plotted from the collected motion capture flight data. 
        The dashed black line represents the reference geometric path, the dashed blue line shows the tracked trajectory generated by TOPPQuad, and the solid red lines show the tracked trajectories output by our model across different runs.
    }
    \label{fig:hardware-0.1}
\vspace{-10pt}
\end{figure}

\begin{figure}[h]
    \centering
    \includegraphics[width=0.95\linewidth]{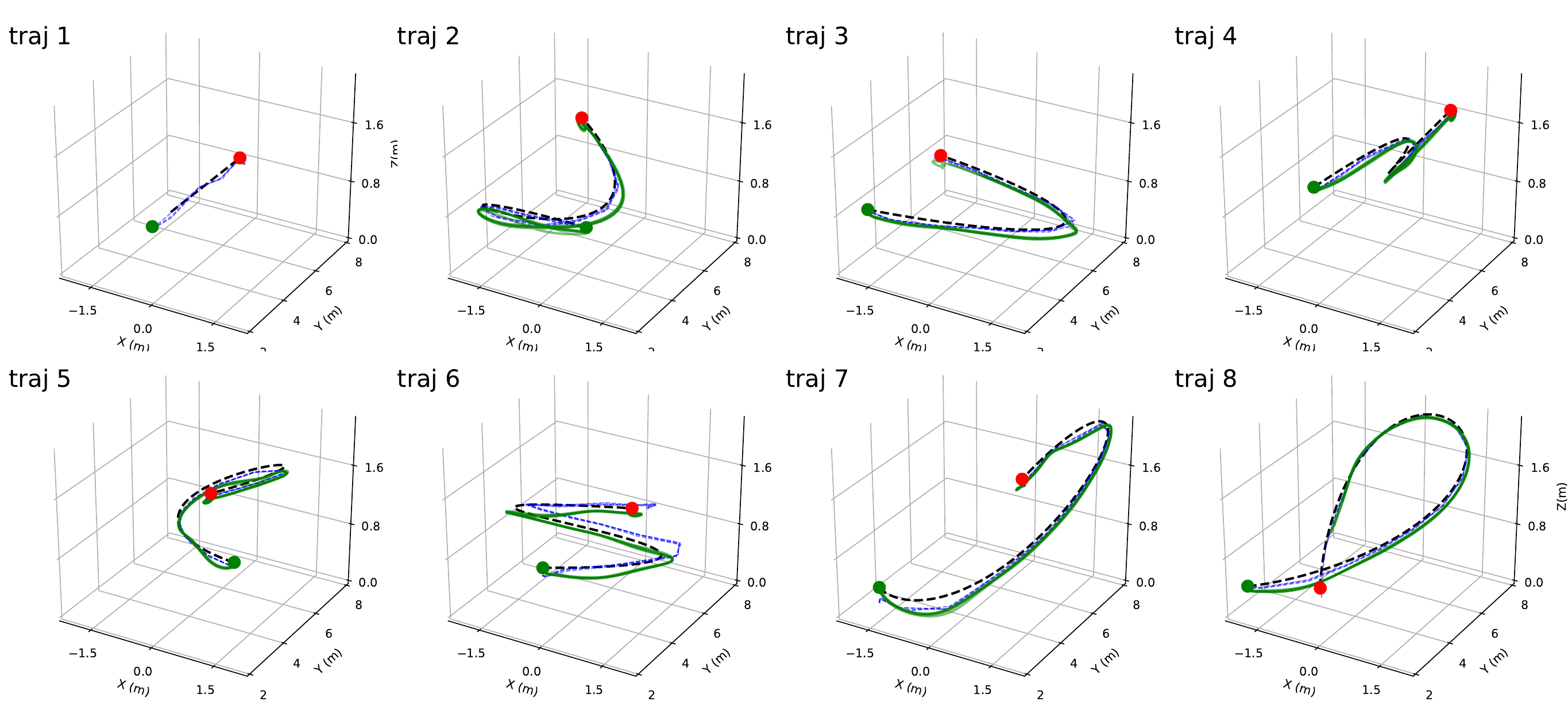}
    \caption{
        \small Experiment visualization of trajectories predicted from the \textbf{LSTM-0.01} model, plotted from the collected motion capture flight data. 
        The dashed black line represents the reference geometric path, the dashed blue line shows the tracked trajectory generated by TOPPQuad, and the solid green lines show the tracked trajectories output by our model across different runs
        All trials of Traj 1 from the predicted model crashed.
    }
    \label{fig:hardware-0.01}
\vspace{-10pt}
\end{figure}

In this section, we present additional hardware experiments with trajectories predicted from the LSTM-0.1 (Fig. \ref{fig:hardware-0.1}) and LSTM-0.01 models (Fig. \ref{fig:hardware-0.01}). Each subfigure represents 5 trials of the learned trajectory and 4 trials of the TOPPQuad trajectory. Table \ref{tab:hardware-noise} compares the performance of these additional models with the LSTM model presented in the main paper. We show all three models have good tracking performance and similar travel times. However, we note that for the \textbf{LSTM-0.01} model, all trials of 'Traj 1' crashed. While the trajectory itself appears very simple, a model trained on a dataset of polynomial paths would be more likely to struggle generalizing to a straight line path.

\begin{table*}[h]
\centering
\small
\setlength{\tabcolsep}{4.5pt} 
\renewcommand{\arraystretch}{1.2} 
\begin{tabular}{c c c c c}
\toprule
 & \textbf{TOPPQuad} &
 \textbf{LSTM} &
 \textbf{LSTM-0.1} &
 \textbf{LSTM-0.01}\\
\midrule
\textbf{max deviation (m)} & 0.347 & 0.355 & 0.373 & 0.372 \\ 
\textbf{travel time (s)} & 7.981 & 8.355 & 8.288 & 8.114\\  
\bottomrule
\end{tabular}
\caption{\small Hardware experiments.
}
\label{tab:hardware-noise}
\vspace{-10pt}
\end{table*}

\newpage


\bibliography{supp_references}  